\documentclass[Journal,InsideFigs,letterpaper,NoLineNumbers]{ascelike-new}
\usepackage[utf8]{inputenc}
\usepackage[T1]{fontenc}
\usepackage{lmodern}
\usepackage{graphicx}
\usepackage[style=base,figurename=Fig.,labelfont=bf,labelsep=period]{caption}
\usepackage{subcaption}
\usepackage{amsmath}

\usepackage{newtxtext,newtxmath}
\usepackage[colorlinks=true,citecolor=black,linkcolor=black]{hyperref}
\usepackage{adjustbox}
\usepackage{multirow}
\usepackage{threeparttable}

\begin{document}

\title{Can AI Master Construction Management (CM)? Benchmarking State-of-the-Art Large Language Models on CM Certification Exams}

\author[1]{Ruoxin Xiong, Ph.D., Aff.M.ASCE}
\author[2]{Yanyu Wang, Ph.D., A.M.ASCE}
\author[3]{Suat Gunhan, Ph.D., M.ASCE}
\author[4]{Yimin Zhu, Ph.D., A.M.ASCE}
\author[5]{Charles Berryman, Ph.D.}

\affil[1]{Assistant Professor, Construction Management Program, College of Architecture \& Environmental Design, Kent State University, Kent, OH, 44242, United States. Email: rxiong3@kent.edu}

\affil[2]{Assistant Professor, Bert S. Turner Department of Construction Management, Louisiana State University, Baton Rouge, LA, 70803, United States (corresponding author). Email: yanyuwang@lsu.edu}

\affil[3]{Professor, Construction Management Program, College of Architecture \& Environmental Design, Kent State University, Kent, OH, 44242, United States. Email: sgunhan@kent.edu}

\affil[4]{Professor, Bert S. Turner Department of Construction Management, Louisiana State University, Baton Rouge, LA, 70803, United States. Email: yiminzhu@lsu.edu}

\affil[5]{Professor, Bert S. Turner Department of Construction Management, Louisiana State University, Baton Rouge, LA, 70803, United States. Email: cberryman@lsu.edu}

\maketitle

\begin{abstract}
The growing complexity of construction management (CM) projects, coupled with challenges such as strict regulatory requirements and labor shortages, requires specialized analytical tools that streamline project workflow and enhance performance. Although large language models (LLMs) have demonstrated exceptional performance in general reasoning tasks, their effectiveness in tackling CM-specific challenges, such as precise quantitative analysis and regulatory interpretation, remains inadequately explored. To bridge this gap, this study introduces \textsc{CMExamSet}, a comprehensive benchmarking dataset comprising 689 authentic multiple-choice questions sourced from four nationally accredited CM certification exams. Our zero-shot evaluation assesses overall accuracy, subject areas (e.g., construction safety), reasoning complexity (single-step and multi-step), and question formats (text-only, figure-referenced, and table-referenced). The results indicate that GPT-4o and Claude~3.7 surpass typical human pass thresholds (70\%), with average accuracies of 82\% and 83\%, respectively. Additionally, both models performed better on single-step tasks, with accuracies of 85.7\% (GPT-4o) and 86.7\% (Claude 3.7). Multi-step tasks were more challenging, reducing performance to 76.5\% and 77.6\%, respectively. Furthermore, both LLMs show significant limitations on figure-referenced questions, with accuracies dropping to approximately 40\%. Our error pattern analysis further reveals that conceptual misunderstandings are the most common (44.4\% and 47.9\%), underscoring the need for enhanced domain-specific reasoning models. These findings underscore the potential of LLMs as valuable supplementary analytical tools in CM, while highlighting the need for domain-specific refinements and sustained human oversight in complex decision making. 
\end{abstract}

\section{Introduction}
The construction industry is undergoing a transformation driven by digital technologies, increased project complexity, heterogeneous regulations, and ongoing labor shortages \cite{abioye2021artificial}. These changes create a pressing need for intelligent tools that can augment human expertise and support decision-making in construction management (CM) \cite{regona2022opportunities}. Among these technologies, large language models (LLMs) such as GPT-4 and Claude have shown a comparative performance in general reasoning, natural language understanding, and educational applications \cite{ooi2025potential}. Their ability to process complex inputs and provide context-aware outputs suggests promising applications in multiple phases of construction projects \cite{regona2022opportunities}.

Despite the recognition of LLM potential, empirical studies evaluating their applicability and reliability in various CM-specific tasks, such as safety analysis, cost estimation, and project scheduling, remain insufficiently explored \cite{sammour2024responsible,barcaui2023better}. These tasks often require complex reasoning, precise numerical analysis, and the interpretation of multimodal information (e.g., technical tables, text, and drawings) \cite{ahmed2014key}. These tasks also involve various domain-specific expertise, ethical considerations, and practical decision making under uncertainty, which requires a rigorous benchmark of LLM capabilities in different CM scenarios \cite{xiong2024transforming}. Benchmarks offer a consistent and structured way to evaluate model performance in representative tasks and scenarios \cite{drori2023dataset}. However, existing LLM benchmarks are largely designed for general-purpose or applications and do not reflect the specialized demands of CM workflows \cite{hendrycks2020measuring,drori2023dataset}. Without a systemic benchmark tailored to construction-specific challenges, the potential of LLMs in effectively supporting professional CM workflows and decision-making processes remains unclear.

To address these gaps, this study introduces \textsc{CMExamSet}, a curated dataset comprising 689 multiple-choice questions (MCQs) sourced from four major CM certification exams: Certified Associate Constructor (CAC), Certified Professional Constructor (CPC), Certified Associate Construction Manager (CACM), and Certified Construction Manager (CCM) \cite{AIC_StudyGuide_2022,AIC_2024_CPC_StudyGuide,CMAA_CACM_Study_Guide,CMAA_CCM_Study_Guide}. Using a zero-shot evaluation approach, we systematically assess the performance of state-of-the-art LLMs, including GPT-4o and Claude 3.7, in overall accuracy, reasoning complexities (single-step and multi-steps) and question formats (text-only, figure-referenced, and table-referenced). This study also analyzes performance by various subject areas (e.g., construction safety, scheduling, and estimating) and classifies the error patterns (e.g., concept misunderstanding and reading or interpretation errors) presented by LLMs, which guides the analysis of LLM limitations in CM practice.

The research questions guiding this study include: (1) How do state-of-the-art LLMs perform in CM core knowledge areas? (2) What specific task formats and reasoning complexities present challenges or advantages for these models? (3) What implications and limitations do the findings suggest regarding the practical adoption of LLMs in construction education and practice? 

This study makes three primary contributions: First, the authors establish a benchmarking framework in CM using standardized practice questions derived from nationally accredited certification exams. Second, this study performs a systematic performance comparison of state-of-the-art LLMs in diverse CM knowledge areas and task complexities, providing a structured assessment of their strengths and limitations. Third, we provide insights into the educational, ethical, and operational implications of integrating advanced LLM technologies into CM practice.

\section{Literature Review}
This section provides an overview of the current literature on LLMs, their applications in CM, existing benchmarking frameworks, and the research gaps that the present study aims to address. 

\subsection{Capabilities of Large Language Models and Their Relevance to Construction Management}
LLMs represent a new advance in artificial intelligence (AI). Built on deep neural network architectures and trained on massive datasets, these models are capable of generating and understanding human-like text \cite{ooi2025potential}. These models learn from vast amounts of data, including books, articles, reports, and technical documents, to capture linguistic patterns, context, and even subtle reasoning steps \cite{chang2024survey}. Unlike earlier rule-based or statistical approaches, LLMs such as GPT-4 and Claude can interpret context over long passages, make inferences, and even perform multi-step reasoning \cite{naveed2023comprehensive}. This means that LLMs can, for example, analyze complex regulatory documents, extract relevant technical information, or synthesize data from diverse sources to support decision making \cite{sammour2024responsible}. In CM, where professionals routinely work with technical documents, regulatory codes, and project documentation, the potential of LLMs to process and synthesize diverse textual inputs shows promise \cite{xiong2024transforming}. Their flexibility in handling various textual styles and formats may offer useful support for decision-making processes in this domain.

\subsection{Applications of Large Language Models in Construction Management}
LLMs offer promising applications in the various phases of construction projects by leveraging their advanced capabilities in content processing, generation, and reasoning \cite{regona2022opportunities,xiong2024transforming}. In the pre-construction phase, these models can extract and synthesize key information from contracts, technical specifications, and regulatory documents, thus supporting planning, cost estimation, and scheduling processes \cite{barcaui2023better,wong2024construction}. During construction, LLMs can help in real-time decision making by summarizing safety protocols, incident reports, and compliance guidelines, which improves risk management and overall site coordination \cite{sammour2024responsible,pu2024autorepo}. In the post-construction phase, these models help prepare comprehensive project documentation and performance reports, capturing lessons learned and best practices for future reference \cite{ghimire2023generative,ahmadi2025automatic}. These investigations underscore the potential benefits of LLMs, yet systematic evaluations that replicate the rigor and breadth of real-world CM scenarios remain in the early stages. In particular, few studies have directly tested LLM performance against the comprehensive standards embodied in various CM knowledge domains, such as construction safety, construction estimation, and scheduling. The lack of comprehensive and comparative evaluations limits our understanding of how these models perform in different knowledge areas and conditions that mimic real-world CM challenges.

\subsection{Existing LLM Benchmarks and the Need for Domain-Specific Evaluation}
Benchmarking frameworks play a pivotal role in evaluating the capabilities, reliability, and limitations of LLMs across domains \cite{drori2023dataset}. These frameworks typically involve curated datasets, task definitions, scoring criteria, and performance baselines to enable consistent and replicable evaluation. Table~\ref{tab:dataset_summary} summarizes representative benchmarks used to assess LLM performance in general and science, technology, engineering, and mathematics (STEM) related domains. These include datasets based on academic exams, textbook questions, and online educational resources.

While these benchmarks have contributed significantly to evaluating and understanding LLM performance in domains, they often do not capture the unique complexities of CM professional practice. CM tasks require interpreting technical documents, applying regulatory codes, managing project constraints, and communicating across disciplines, which are not typically represented in existing LLM benchmarks \cite{barcaui2023better,sammour2024responsible,ahmed2014key}. Therefore, a domain-specific evaluation framework is necessary to assess LLMs against the diverse and complex demands of real-world CM practice, ensuring comprehensive data representativeness and coverage of practical scenarios.

\begin{table}[htbp]
\centering
\caption{Benchmarks and datasets for LLM performance evaluation}
\begin{adjustbox}{max width=\textwidth}
\label{tab:dataset_summary}
\begin{tabular}{p{5.5cm} p{3cm} p{2.5cm} p{5cm}}
\hline
\textbf{Dataset} & \textbf{Source} & \textbf{Domain} & \textbf{Data Type} \\
\hline
M3Exam \cite{zhang2023m3exam} & Graduation exams & General & Question \& answer (Q\&A) pairs \\
SciQ \cite{welbl2017crowdsourcing} & Science textbooks & Science & Multiple-choice Q\&A pairs \\
ScienceQA \cite{lu2022learn} & Online learning platform & Science & Multiple-choice Q\&A pairs \\
SciBench \cite{wang2023scibench} & University and course exams & Science & Open-ended questions and step-by-step solutions \\
University STEM Courses Dataset \cite{drori2023dataset} & University course & STEM & Q\&A pairs \\
MMLU \cite{hendrycks2020measuring} & Online resources & Multidomain & Multiple-choice Q\&A pairs \\
\hline
\end{tabular}
\end{adjustbox}
\end{table}

\subsection{Research Gaps and Contributions}
Despite the rapid advancements in LLMs and their demonstrated potential in CM, three main research gaps remain in the context of CM: (1) a lack of CM-specific datasets capturing the complexity of CM tasks such as cost estimation, regulatory compliance, and scheduling; (2) limited comparative analysis across models and scenarios, which systematically evaluate how different LLMs perform across a diverse range of CM-relevant tasks and knowledge areas; and (3) insufficient understanding of the practical implications and limitations associated with integrating LLMs into professional CM practice, such as reasoning limitations and ethical concerns.

To address these gaps, this study makes three primary contributions:
\begin{itemize}
    \item Establishes the comprehensive benchmarking framework, \textsc{CMExamSet}, specifically tailored to CM, built using practice questions from nationally accredited certification exams administered by the American Institute of Constructors (AIC) and the Construction Management Association of America (CMAA).
    \item Provides comparative evaluations of state-of-the-art LLMs in CM, systematically assessing their strengths and limitations across a diverse range of scenarios and subject areas.
    \item Offers insights into the practical, ethical, and educational dimensions of LLM adoption, guiding responsible LLM integration into CM practices and education.
\end{itemize}

\section{\textsc{CMExamSet}: Construction Management Certification Exam Dataset}
\begin{figure}[htbp]
    \centering
    \includegraphics[width=0.98\linewidth]{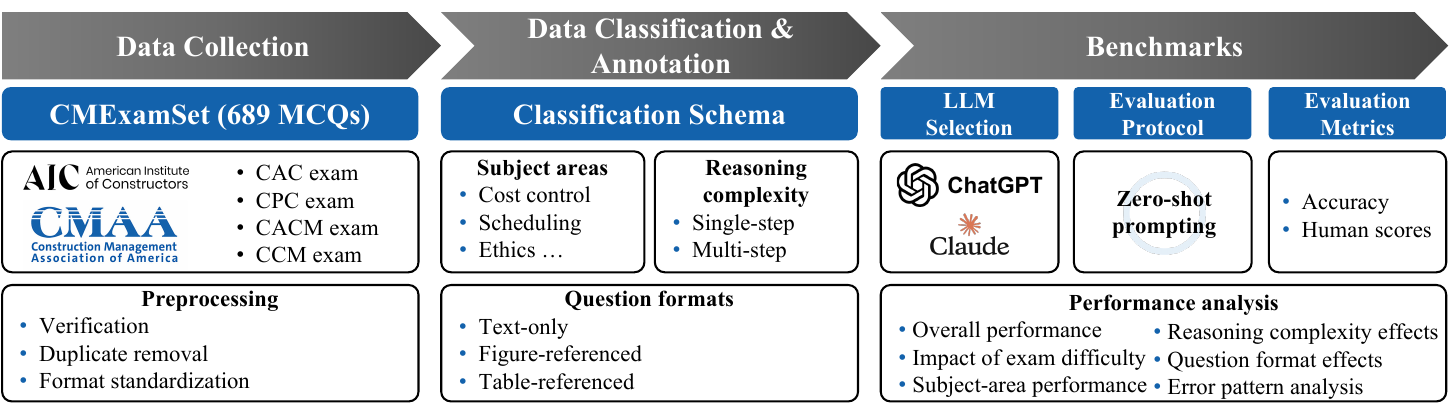}
    \caption{Benchmark framework for LLM performance evaluation in CM}
    \label{fig:framework}
\end{figure}

This study introduces a benchmark framework to evaluate LLM performance in CM, as shown in Figure~\ref{fig:framework}. The framework is built upon a curated dataset of 689 multiple-choice questions extracted from CM certification exams. Each question is subjected to data cleaning, standardization and systematic classification by subject area, reasoning complexity, and format, ensuring alignment with real-world professional standards and a broad representation of CM knowledge areas. Benchmarking is conducted through zero-shot prompting from leading LLMs, with performance assessed through accuracy metrics and detailed error analysis to uncover domain-specific challenges. 

\subsection{Data Collection and Preprocessing}
This study introduces \textsc{CMExamSet}, a curated dataset of CM certification exam questions compiled to benchmark LLM. CM professional certification exams, such as those administered by the AIC and CMAA, are designed to assess comprehensive competencies in key professional domains \cite{AIC_StudyGuide_2022,CMAA2022}. These domains include project scheduling, cost estimation, safety protocols, ethical standards, and contract management. As these certifications are accredited by the ANSI National Accreditation Board (ANAB), they reflect rigorous industry standards and the multifaceted challenges encountered in construction practice \cite{AIC2023}. Evaluating LLM performance on these exams provides a structured and rigorous approach to assessing their capabilities, as success in CM tasks requires general knowledge retrieval and reasoning and the application of domain-specific expertise in complex and context-sensitive decision making \cite{wao2022relationship}.

Table~\ref{tab:cm_exam_dataset} summarizes the data sources of \textsc{CMExamSet}, which covers a wide range of knowledge areas of CM, including cost control, contract administration, safety, and ethics. The dataset consists of MCQs extracted from official study guides corresponding to the CAC, CPC, CACM, and CCM exams. These materials are proprietary and not publicly accessible, minimizing the risk of data leakage for LLM evaluation. In total, 689 MCQs were compiled, each containing a clearly defined question stem, four answer choices (A, B, C, D), and an official answer key. The dataset includes text-based questions as well as those requiring interpretation of figures and tables, ensuring a diverse representation of complexity levels and professional competencies.

\begin{table}[htbp]
\centering
\caption{Data sources of the \textsc{CMExamSet}}
\label{tab:cm_exam_dataset}
\begin{adjustbox}{max width=\textwidth}
\begin{tabular}{p{1cm}p{5.5cm}p{1.5cm}p{2cm}p{3cm}p{3.5cm}}
\hline
Exams & Description & Authority & \# Questions & Question Types & Source\\ \hline
CAC & Level I Construction Fundamentals Examination  & AIC & 100 & MCQs & CAC Study Guide \cite{AIC_StudyGuide_2022} \\

CPC & Level II Advanced Construction Applications Examination & AIC & 489 & MCQs & CPC Study Guide \cite{AIC_2024_CPC_StudyGuide} \\

CACM & Certified Associate Construction Manager Examination & CMAA & 50 & MCQs & CACM Study Guide \cite{CMAA_CACM_Study_Guide} \\

CCM & Certified Construction Manager Examination & CMAA & 50 & MCQs & CCM Study Guide \cite{CMAA_CCM_Study_Guide} \\
\hline
\end{tabular}
\end{adjustbox}
\end{table}

The preprocessing phase involved systematic quality control to ensure the dataset's accuracy, consistency, and applicability for LLM evaluation. These steps include: (i) Verification: Each question was reviewed for accuracy and consistency with the study materials. (ii) Duplication removal: Duplicate questions were identified and removed to prevent redundancy. (iii) Format standardization: The question structures were standardized for comparative analysis, with all references (figures, tables, and supplementary materials) labeled and formatted for consistency.

\subsection{Question Classification and Annotation}
To enable a detailed analysis of LLM performance in CM practices, \textsc{CMExamSet} employs a structured classification system that categorizes each question based on its subject area, reasoning complexity, and question format (Table~\ref{tab:classification-schema}). This classification framework provides insight into the potential of LLMs in different CM competencies and levels of cognitive demand.

\begin{table}[htbp]
\centering
\caption{Classification schema for \textsc{CMExamSet} questions}
\label{tab:classification-schema}
\begin{adjustbox}{max width=\textwidth}
\begin{tabular}{p{2.5cm}p{6.5cm}p{7.5cm}}
\hline
\textbf{Dimension} & \textbf{Description} & \textbf{Example} \\
\hline
Question ID & Unique identifier assigned to each question & CPC-3-Q10 \\

Subject areas & Specific subject area assessed & Cost Control, Risk Management, Ethics \\

Reasoning complexity & Cognitive demand: single-step or multi-step reasoning & \textit{"Q: The owner wants to fast-track a construction project. Which project delivery system best supports this process?"} (single-step) \\

Question format & Format classification: text-only, figure-referenced, table-referenced & \textit{"Q: Using the information provided in \textbf{\textsc{Figure 10}}, what are the critical activities for this logic network?"} (figure-referenced) \\

Source reference & Original source document of the question & 2024 CPC Study Guide, Chapter \#3, Question 10 \\
\hline
\end{tabular}
\end{adjustbox}
\end{table}

The subject area classification spans various CM topics such as contract administration, cost control, scheduling, safety management, and ethics, ensuring comprehensive coverage of industry-relevant knowledge areas. The reasoning complexity metric differentiates \textit{single-step questions}, which require straightforward concept recall or direct application, from \textit{multi-step questions}, which require integrative reasoning, numerical analysis, or cross-referencing multiple sources.

The question format dimension distinguishes between \textit{text-only} questions and those requiring reference to \textit{figures} and \textit{tables}, reflecting real-world construction scenarios where professionals interpret technical documentation such as drawings, site plans, engineering schematics, and tabular data. The source reference records the exam materials from which each question was derived.

\subsection{Dataset Characteristics}
The \textsc{CMExamSet} dataset comprises 689 MCQs, covering various CM domains and cognitive demand levels. Tables~\ref{tab:content-areas} and~\ref{tab:characteristics} summarize the data sources, detailing question distributions in subject areas, reasoning complexity, and question formats.

\begin{table}[htbp]
\centering
\caption{Distribution of questions by subject areas in \textsc{CMExamSet}}
\label{tab:content-areas}
\begin{adjustbox}{max width=\textwidth}
\begin{tabular}{p{1cm}p{11cm}p{5cm}}
\hline
\textbf{Exams} & \textbf{Subject areas (\# Questions)} & \textbf{Examples in subject areas} \\
\hline
CAC & Communication skills (11), Engineering concepts (8), Management concepts (10), Materials, methods \& project modeling (10), Bidding \& estimating (10), Budgeting, costs \& cost control (11), Planning, scheduling \& schedule control (11), Construction safety (10), Construction geomatics (8), Project administration (11) & \textit{\textbf{Bidding \& estimating:}} procurement and bidding process, estimates, quantity takeoff\\
CPC & Project scope development (98), Employment practices (23), Working relationships (66), Start-up \& support (13), Resource management (87), Cost control (71), Project closeout (10), Safety management (88), Ethics (7), Contract interpretations (26) & \textit{\textbf{Ethics:}} business ethics, professional practice ethics \\
CACM & Program \& project management (5), Cost management (5), Time management (5), Quality management (5), Contract administration (5), Safety management (5), Risk management (5), Professional practice (5), Sustainability (5), Technology (5) & \textit{\textbf{Risk management:}} identify risk, evaluate risk, risk monitoring, change orders, etc. \\
CCM & Program \& project management (5), Cost management (5), Time management (5), Quality management (5), Contract administration (5), Safety management (5), Risk management (5), Professional practice (5), Sustainability (5), Technology (5) & \textit{\textbf{Technology:}} BIM/VDC model, emerging technologies, project data, etc. \\
\hline
\end{tabular}
\end{adjustbox}
\end{table}

\begin{table}[htbp]
\caption{Reasoning complexity and question format distribution in \textsc{CMExamSet}}
\label{tab:characteristics}
\begin{adjustbox}{max width=\textwidth}
\begin{tabular}{cccccccc}
\hline
\multirow{2}{*}{\textbf{Exams}} &
  \multirow{2}{*}{\textbf{\# Questions}} &
  \multicolumn{2}{c}{\textbf{Reasoning complexity}} &
  \textbf{} &
  \multicolumn{3}{c}{\textbf{Question formats}} \\ \cline{3-4} \cline{6-8} 
    &     & Single-step & Multi-step &  & Text-only & Table-referenced & Figure-referenced \\ \hline 
CAC                & 100 & 68 & 32 &  & 72 & 21 & 7 \\
CPC                & 489 & 317 & 172 & & 368 & 121 & 0 \\
CACM               & 50 & 24 & 26 &  &  50  & 0  &  0 \\
CCM                & 50  & 4  & 46  &  & 47 & 1  & 2 \\ \hline
\textbf{Total}     & 689 & 412 & 277 &  & 537 & 143 & 9 \\
\textbf{Ratio (\%)} & -- & 59.8 & 40.2 &  & 77.9 & 20.8  & 1.3 \\ \hline
\end{tabular}
\end{adjustbox}
\end{table}

\subsubsection{Subject Area Coverage}
As shown in Table~\ref{tab:content-areas}, \textsc{CMExamSet} captures the full spectrum of CM knowledge in nationally accredited certification exams. The dataset includes topics ranging from cost estimation and project scheduling to contract administration, construction safety, and risk management. Across all certification exams, the dataset ensures balanced coverage of critical CM knowledge areas, supporting a comprehensive LLM evaluation framework.

\subsubsection{Reasoning Complexity and Question Formats}
As illustrated in Table \ref{tab:characteristics}, about 60\% of the questions in \textsc{CMExamSet} require a single-step reasoning, assessing fundamental CM knowledge through direct recall or straightforward application. The remaining 40\% involve multi-step reasoning, requiring interpretation, computation, or the integration of multiple data sources to arrive at a correct response. 

In terms of question format, \textsc{CMExamSet} reflects a diverse range of CM evaluation scenarios: 77.9\% of the questions are text-only, testing conceptual understanding and theoretical knowledge in CM. 20.8\% of the questions require table interpretation, evaluating numerical reasoning, data analysis, and the ability to extract insights from structured information. 1.3\% of the questions involve figure references, integrating tasks such as reading construction drawings, interpreting schematic diagrams, or analyzing graphical representations of project workflows.

\section{Benchmarking Large Language Models for Construction Management}
This section details the methodology for evaluating state-of-the-art LLMs on \textsc{CMExamSet}. The evaluation framework includes LLM selection, experimental setup, performance metrics, and a comparison with human performance to contextualize model results against certification standards.

\subsection{Model Selection and Experimental Framework}

\subsubsection{Selection of LLMs}
This study focuses on evaluating GPT-4o \cite{OpenAI2024} and Claude 3.7 \cite{Anthropic2025}, two high-performing and widely accessible LLMs, to assess their capabilities in solving CM questions. These models have demonstrated strong performance in structured reasoning, generalization, and knowledge-intensive tasks across multiple domains \cite{huang2024olympicarena,myrzakhan2024open}, making them well-suited to evaluate AI proficiency in professional assessments.

The selection of GPT-4o and Claude 3.7 was informed by their maturity, stability of access, and consistent performance in comparative evaluations \cite{wu2023comparative,kevian2024capabilities}. GPT-4o has shown superior performance in tasks requiring comprehension, logical reasoning, and problem-solving \cite{OpenAI2024}. Claude 3.7 has demonstrated competitive accuracy in tasks involving structured data analysis and decision-making \cite{Anthropic2025}. Although other contemporary LLMs, such as Gemini \cite{google2023gemini} and Llama \cite{touvron2023llama}, also show potential, the selection of GPT-4o and Claude 3.7 provides a practical and controlled foundation for a focused analysis of LLM capabilities within the specific demands of CM tasks.

\subsubsection{Zero-Shot Evaluation Protocol and Prompting Strategy}
The evaluation follows a zero-shot prompting protocol \cite{brown2020language,liusie2023llm}, in which models are tested without any fine-tuning, instructions, examples, or domain-specific training materials. This design avoids introducing external biases or prompt engineering effects, allowing the evaluation of the models’ inherent reasoning capabilities and their ability to generalize to CM tasks \cite{liu2023pre}. This practice is consistent with standard practices in the foundational LLM benchmarks, such as MMLU \cite{hendrycks2020measuring} and M3Exam \cite{zhang2023m3exam}.

To ensure a fair and unbiased assessment, we adopted a zero-shot setup: Each model was presented with the original MCQ content of \textsc{CMExamSet}, exactly as it would appear to a human test-taker. Specifically, the prompt included: (1) the full text of each question, (2) four answer options (A, B, C, D), and (3) any accompanying figures or tables referenced in the question. Each LLM model completed the entire \textsc{CMExamSet}, and the responses were recorded without manual corrections, filtering, or post-processing.

\subsection{Evaluation Metrics}
Following established practices in LLM benchmarking studies \cite{hendrycks2020measuring,zhang2023m3exam,lu2022learn}, model performance was evaluated using accuracy, defined as the percentage of correctly answered questions based on official answer keys. Accuracy assesses how well LLMs perform on professional certification-style MCQs.

To capture performance from multiple perspectives, this study reports three levels of accuracy: (1) \textit{Overall accuracy} represents the percentage of correct responses to all the questions in \textsc{CMExamSet}. (2) \textit{Subject-specific accuracy} assesses model performance within individual knowledge domains such as cost control, contract administration, and safety management. (3) \textit{Question-type accuracy} categorizes results based on question format, including text-only, table-referenced, and figure-referenced questions, as well as reasoning complexity, including single-step and multi-step.

\subsection{Human Performance}
To contextualize LLM performance, the results are compared with the passing thresholds for the CM certification exams. For AIC-administered exams, including CAC and CPC certifications, a passing threshold of 70\% is commonly used. Achieving or surpassing this benchmark indicates performance comparable to that of a minimally qualified human candidate. For CMAA-administered exams, including the CACM and CCM certifications, there is no fixed passing score. Instead, the passing threshold varies in different versions of the exam to account for variations in test difficulty.

Further grounding is provided by recent exam statistics: the Spring 2024 CAC exam reported a pass rate of approximately 34\% (233 out of 691 candidates), highlighting the real-world difficulty of CM certification exams. While not a direct performance benchmark, these data offer insight into the distribution of human outcomes and help interpret the relative performance of LLMs.

\section{Results and Analysis}
This section presents the evaluation results of two LLMs, GPT-4o and Claude 3.7, on \textsc{CMExamSet}. The analysis examines model accuracy, performance across certification levels, subject-area performance, reasoning complexity, question format effects, and common error patterns.

\subsection{Overall Performance Across Certification Exams}
\begin{figure}[htbp]
    \centering
    \includegraphics[width=0.98\linewidth]{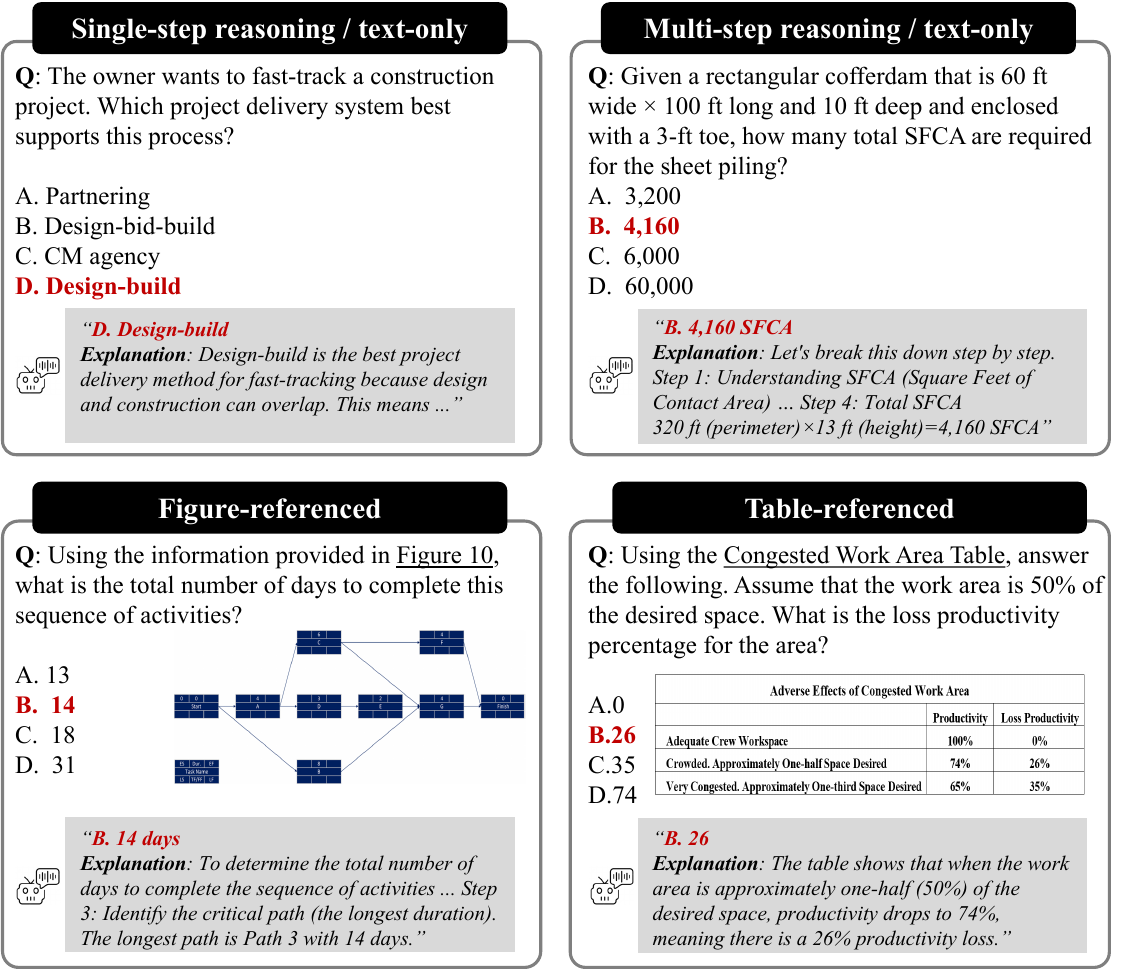}
    \caption{Sample correct responses from GPT-4o, illustrating how the model handles different reasoning complexity and question formats.}
    \label{fig:questions}
\end{figure}

The questions vary in complexity, covering a range of single-step reasoning tasks, multi-step reasoning problems, figure-referenced questions, and table-referenced questions. Fig.~\ref{fig:questions} illustrates how LLMs successfully handle different types of reasoning and question formats. Single-step reasoning questions require direct knowledge recall (e.g., identifying design-build as the best fast-tracking method), while multi-step reasoning questions involve step-by-step calculations (e.g., computing 4,160 SFCA for sheet piling). Figure-referenced questions require interpreting graphical information (e.g., determining 14 days from a precedence diagram), whereas table-referenced questions involve extracting data (e.g., identifying 26\% productivity loss from a workspace table). These categories highlight varying levels of complexity in assessing construction knowledge.

\begin{figure}[htbp]
    \centering
    \includegraphics[width=0.95\linewidth]{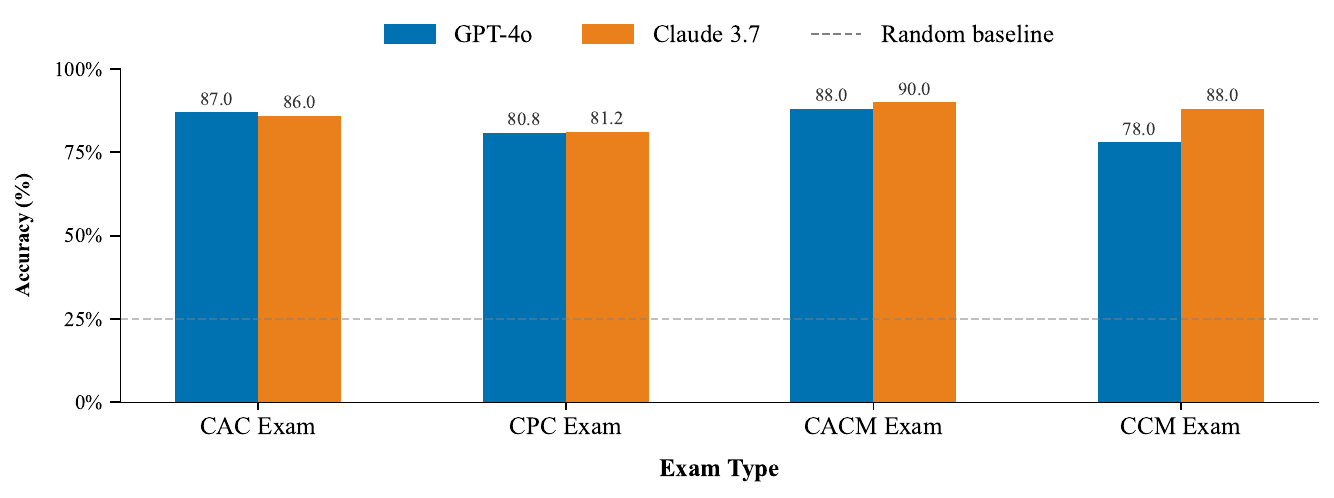}
    \caption{Exam-level performance comparison on \textsc{CMExamSet}}
    \label{fig:score}
\end{figure}

Fig.~\ref{fig:score} compares the normalized scores of GPT-4o and Claude 3.7 on four exams: CAC, CPC, CACM, and CAM. Both LLM models consistently outperform random baselines. In particular, GPT-4o achieves higher accuracy in CAC (87.0\%), while Claude 3.7 outperforms in CPC (81.2\%), CACM (90.0\%), and CCM (88.0\%).

To evaluate whether the differences in exam-level accuracy between the GPT-4o and Claude models are statistically significant, we conducted a paired two-tailed \emph{t}-test \cite{ross2017paired}:
\begin{equation}
    H_0: \mu_{\text{GPT-4o}} = \mu_{\text{Claude}}, \quad 
H_a: \mu_{\text{GPT-4o}} \neq \mu_{\text{Claude}},
\end{equation}
where \(\mu_{\text{GPT-4o}}\) and \(\mu_{\text{Claude}}\) represent the mean accuracy scores of the GPT-4o and Claude models, respectively, across the four exams.

The test result \(p\)-value (\(p = 0.33\)) indicates that there are no statistically significant differences in the overall performance at the exam level between the two models. Both models exceed the 70\% benchmark, the referenced AIC standard for minimal competence in professional competence.

\subsection{Impact of Exam Difficulty}
Each exam corresponds to a different difficulty level, ranging from foundational (CAC and CACM) and advanced (CPC and CCM). As shown in Table~\ref{tab:exam-difficulty}, GPT-4o's accuracy declines from 87.0\% (CAC) to 80.8\% (CPC) and from 88.0\% (CACM) to 78.0\% (CCM), with Claude 3.7 showing a similar drop. These trends highlight the increased difficulty of professional-level content. Future work should validate these patterns using parallel exam formats or case-based evaluations to better understand LLMs’ sensitivity to domain complexity.

\begin{table}[htbp]
\centering
\caption{Exam-level accuracies (\%) for LLMs}
\label{tab:exam-difficulty}
\begin{adjustbox}{max width=\textwidth}
\begin{tabular}{lcccc}
\hline
\textbf{Exam (Authority)} & \textbf{Difficulty} & \textbf{GPT-4o} & \textbf{Claude 3.7} & \textbf{Diff.\ (GPT-4o $-$ Claude 3.7)}\\
\hline
CAC (AIC)   & Level~I (Foundational) & 87.0 & 86.0 & $+1.0$ \\
CPC (AIC)   & Level~II (Advanced) & 80.8 & 81.2 & $-0.4$ \\ [6pt]
CACM (CMAA) & Associate               & 88.0 & 90.0 & $-2.0$ \\
CCM (CMAA)  & Professional            & 78.0 & 88.0 & $-10.0$ \\
\hline
Average & -- & 82.0 & 83.0 & $-1.0$ \\
\hline
\end{tabular}
\end{adjustbox}
\end{table}

\subsection{Subject-Area Performance}
\begin{figure}[ht]
    \centering
    \begin{subfigure}{0.49\textwidth}
        \centering
        \includegraphics[width=\linewidth]{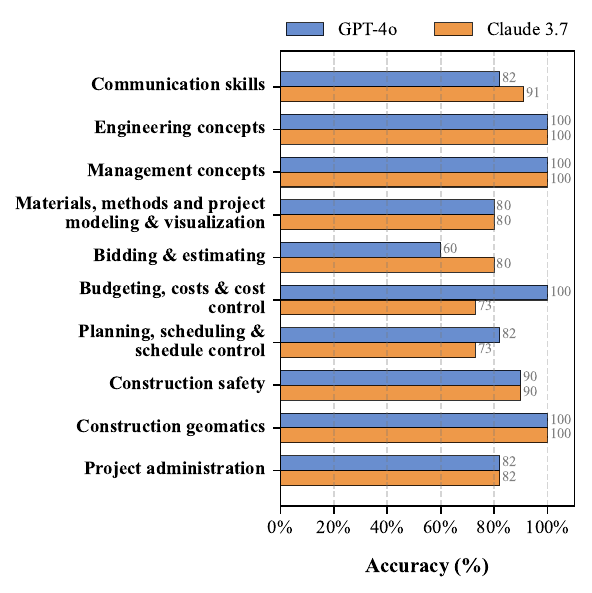}
        \caption{CAC Exams}
        \label{fig:cac_accuracy}
    \end{subfigure}
    \hfill
    \begin{subfigure}{0.49\textwidth}
        \centering
        \includegraphics[width=\linewidth]{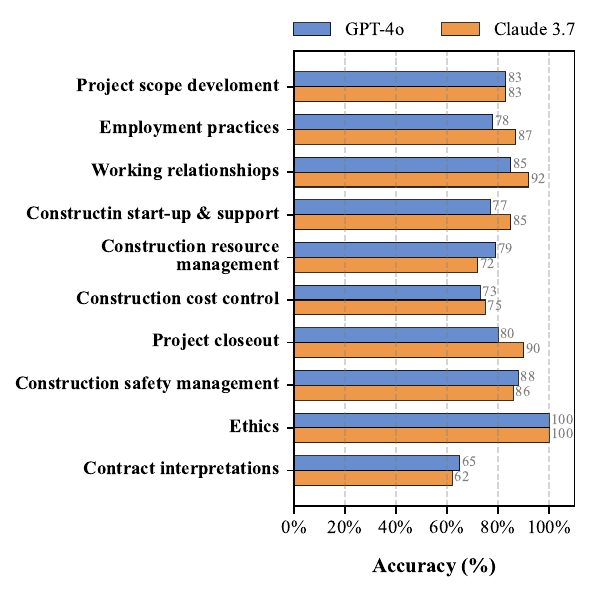}
        \caption{CPC Exams}
        \label{fig:cpc_accuracy}
    \end{subfigure}
    \hfill
    \begin{subfigure}{0.49\textwidth}
        \centering
        \includegraphics[width=\linewidth]{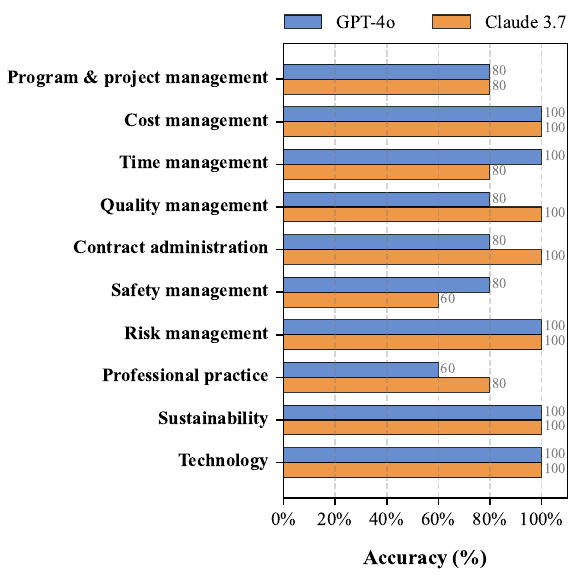}
        \caption{CACM Exams}
        \label{fig:cacm_acuracy}
    \end{subfigure}
    \hfill
    \begin{subfigure}{0.49\textwidth}
        \centering
        \includegraphics[width=\linewidth]{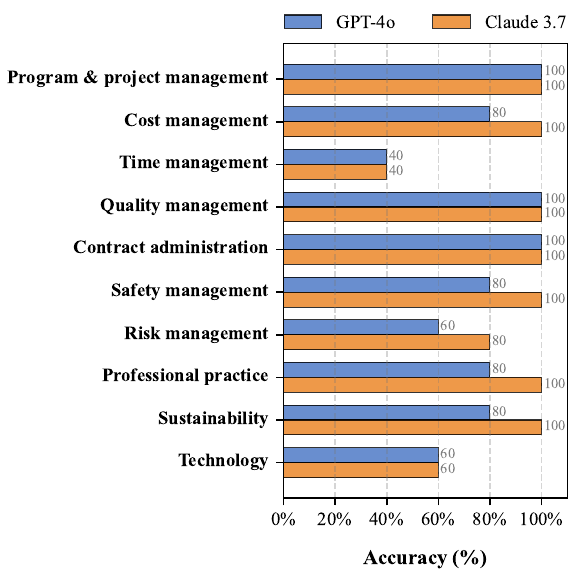}
        \caption{CCM Exams}
        \label{fig:ccm_acuracy}
    \end{subfigure}
    \caption{LLM accuracy (\%) by subject area on \textsc{CMExamSet}}
    \label{fig:content_areas}
\end{figure}

Fig.~\ref{fig:content_areas} illustrates the model accuracy in various knowledge domains, including scheduling, cost control, and safety. Although both models exhibit generally consistent performance across most subject areas, there are variations in specific domains. For example, GPT-4o excels in budgeting and cost control (Fig.~\ref{fig:cac_accuracy}), while Claude 3.7 shows higher accuracy in employment practices and project closeout (Fig. \ref{fig:cpc_accuracy}). However, both models show lower accuracy in some complex reasoning or numerical tasks, such as time management (Fig.~\ref{fig:ccm_acuracy}) and contract interpretation (Fig.~\ref{fig:cpc_accuracy}).

A Wilcoxon signed-rank Test \cite{woolson2005wilcoxon} was performed to assess whether these differences in the subject area were statistically significant. However, there are no statistically significant differences (\(p>0.05\)) in accuracy between GPT-4o and Claude~3.7, indicating that neither model is consistently superior in all domains in \textsc{CMExamSet}. These results suggest that while LLMs effectively handle conceptual knowledge basics, they may face challenges in domains requiring complex calculations, multi-step reasoning, or context-dependent decision-making.

\subsection{Reasoning Complexity Effects}
\label{subsec:reasoning_complexity}
\begin{figure}[htbp]
    \centering
    \includegraphics[width=0.98\linewidth]{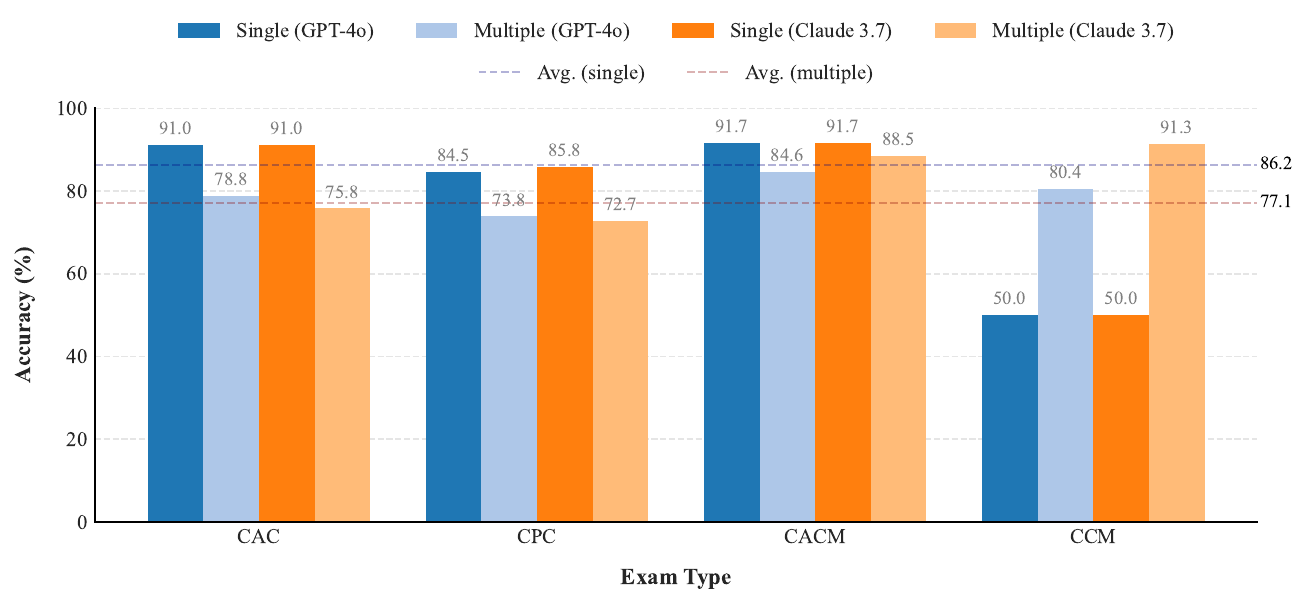}
    \caption{LLM performance on single-step vs.\ multi-step questions on \textsc{CMExamSet}}
    \label{fig:reason}
\end{figure}

Fig.~\ref{fig:reason} compares model accuracy on single-step and multi-step reasoning tasks in four CM exam types. On average, GPT-4o achieved 85.7\% accuracy on single-step questions and 76.5\% on multi-step questions, while Claude 3.7 attained 86.7\% and 77.6\%, respectively. Both models demonstrated stronger performance on single-step tasks, with GPT-4o reaching over 91\% accuracy in CAC and CACM. In contrast, multi-step questions consistently posed greater difficulty for both models, with noticeable declines in accuracy.

To evaluate statistical differences in accuracy between reasoning types, we used a two-proportion \emph{z}-test \cite{lachenbruch2001comparisons} for each model:
\begin{equation}
    H_0: p_{\text{single}} = p_{\text{multi}}, \quad 
H_a: p_{\text{single}} \neq p_{\text{multi}}.
\end{equation}

Table~\ref{tab:statistical} presents the accuracies and \(p\)-values for single-step and multi-step reasoning questions. The results indicate that both models perform better on single-step questions than on multi-step ones (\(p<0.01\)), reflecting the challenge posed by tasks requiring multi-stage logic or computation. These findings underscore the need for improved multi-step reasoning frameworks in CM, where decision-making often requires integrating multiple constraints and iterative problem-solving.

\begin{table}[htbp]
\centering
\begin{threeparttable}
\caption{Two-proportion \emph{z}-tests for single-step vs.\ multi-step questions}
\label{tab:statistical}
\begin{tabular}{lcccc}
\hline
\textbf{Model} & \textbf{Acc. (Single-step) \%} & \textbf{Acc. (Multi-step) \%} & \textbf{\emph{z}} & \textbf{\emph{p}-value}\\
\hline
GPT-4o & 85.7 & 76.5 & 3.06 & 0.002** \\
Claude 3.7 & 86.7 & 77.6 & 3.10 & 0.002** \\
\hline
\end{tabular}
\begin{tablenotes}
\footnotesize
\item \textit{Note.} **$p<0.01$; *$p<0.05$.
\end{tablenotes}
\end{threeparttable}
\end{table}

\subsection{Question Format Effects}
\begin{figure}[htbp]
    \centering
    \includegraphics[width=0.98\linewidth]{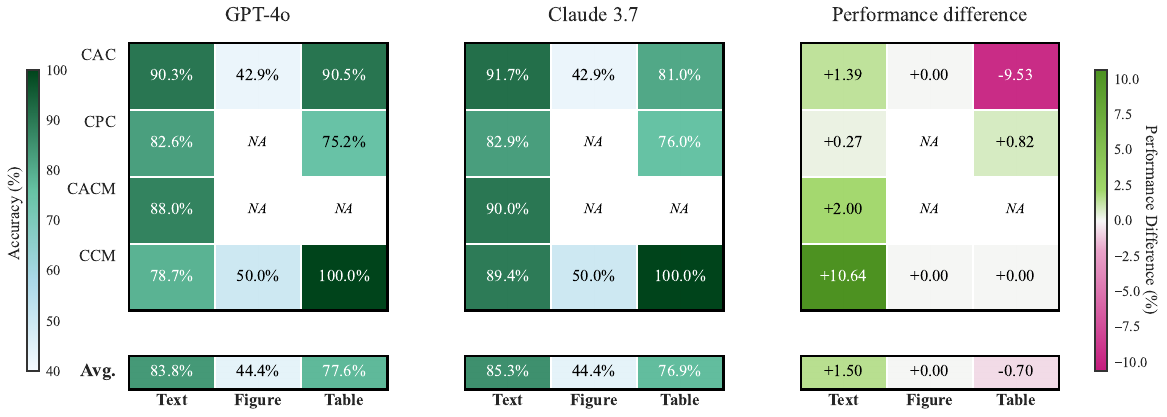}
    \caption{Comparison of model performance by question format on \textsc{CMExamSet}. ``\textit{NA}'' indicates content types not present in particular categories. Performance difference = Claude~3.7 $-$ GPT-4o.}
    \label{fig:format}
\end{figure}

Figure~\ref{fig:format} and Table~\ref{tab:pairwise} illustrate model performance in three question formats: text-based, figure-referenced, and table-referenced. GPT-4o and Claude 3.7 achieve high accuracy in text-based and table-referenced questions, but worse on figure-referenced items.

To assess whether model performance differs by question formats, we conducted a Chi-square test of independence \cite{mchugh2013chi} in different formats. The result ($p<0.01$) indicates a statistically significant relationship between the question format and model accuracy.

To identify specific pairwise differences between formats, we performed two-proportion \emph{z}-tests using pooled data from all available questions (Table~\ref{tab:pairwise}). The results show that both models perform significantly better on text-based questions than figure-referenced ones ($p<0.01$), and also significantly better on table-based than figure-referenced questions (\(p<0.05\)). Claude 3.7 also shows higher accuracy on text-based versus table-referenced questions (\(p<0.05\)).

These results suggest that two LLMs can effectively process textual and structured tabular data, but remain limited in processing visual content such as diagrams. Figure-referenced questions often require recognizing relationships between visual elements, extracting quantitative or qualitative insights, and integrating visual information into reasoning. Given the critical role of visual information in CM, enhancing LLMs with vision-language integration and domain-specific spatial reasoning is essential to improve their applicability in real-world tasks.

\begin{table}[htbp]
\centering
\begin{threeparttable}
\caption{Pairwise two-proportion \emph{z}-tests for question formats}
\label{tab:pairwise}
\begin{tabular}{llcccccc}
\hline
\textbf{Model} & \textbf{Comparison} & 
$n_1$ & \textbf{Acc$_1$ (\%)} &
$n_2$ & \textbf{Acc$_2$ (\%)} &
$\emph{z}$ & \textbf{\emph{p}}\\
\hline
\multirow{3}{*}{GPT-4o}
 & Text vs.\ Figure & 537 & 83.8 & 9   & 44.4 & 3.13 & 0.002** \\
 & Text vs.\ Table  & 537 & 83.8 & 143 & 77.6 & 1.73 & 0.084  \\
 & Figure vs.\ Table & 9   & 44.4 & 143 & 77.6 & -2.25 & 0.025* \\[6pt]
\multirow{3}{*}{Claude 3.7}
 & Text vs.\ Figure & 537 & 85.3 & 9   & 44.4 & 3.37 & 0.0008** \\
 & Text vs.\ Table  & 537 & 85.3 & 143 & 76.9 & 2.40 & 0.017*  \\
 & Figure vs.\ Table & 9   & 44.4 & 143 & 76.9 & -2.18 & 0.029* \\
\hline
\end{tabular}
\begin{tablenotes}
\footnotesize
\item \textit{Note.} *$p < 0.05$, **$p < 0.01$. Acc$_1$ and Acc$_2$ represent the accuracy percentages of the first and second question formats in each comparison, respectively.
\end{tablenotes}
\end{threeparttable}
\end{table}

\subsection{Error Pattern Analysis}
To better understand model limitations in CM problem-solving tasks, we analyzed error patterns in incorrect responses. Motivated by the cognitive processes involved in problem solving \cite{surif2012conceptual}, this study categorized model errors into three main types, (1) type 1: reading or interpretation errors, (2) type 2: conceptual misunderstandings, and (3) type 3: procedural or methodological errors, to diagnose model limitations in CM practices.

\begin{itemize}
    \item \textbf{Type 1: Reading or Interpretation Errors} occur when an LLM incorrectly extracts, associates, or processes textual, numerical, or graphical information. Examples include misinterpreting figures, misreading numerical values, or overlooking critical constraints explicitly stated in the questions.
    \item \textbf{Type 2: Conceptual Misunderstandings} stem from an incomplete or incorrect understanding of the domain-specific principles of LLMs. These errors suggest that the model may not effectively distinguish between similar concepts, which can lead to incorrect definitions, the mixing of distinct terminologies, or the misapplication of theoretical principles.
    \item \textbf{Type 3: Procedural or Methodological Errors} arise when LLMs correctly identify the relevant computational framework or the logical reasoning process but apply incorrect methods or calculations. These errors typically occur in multi-step problem-solving tasks requiring sequential calculations, logical deductions, or rule-based processing.
\end{itemize}

Table~\ref{tab:gpt4_errors} shows examples for each error type. Type 1 errors occur when the model misinterprets information, as seen when GPT-4o incorrectly identifies a contractor’s requirement instead of recognizing structural constraints. Type 2 errors involve incorrect conceptual reasoning, such as Claude 3.7 selecting an incorrect safety meeting term despite understanding its monthly occurrence. Type 3 errors arise from procedural errors in the calculations, exemplified by GPT-4o misapplying a bond premium schedule. These cases illustrate LLMs' challenges in precise reading, domain-specific terminology, and complex problem solving.

\begin{table}[htbp]
\centering
\caption{Examples of error types and explanations in LLM responses}
\label{tab:gpt4_errors}
\begin{adjustbox}{max width=\textwidth}
\begin{tabular}{p{2cm} p{8cm} p{7cm} p{4cm}}
\hline
\textbf{Error Type} & \textbf{Question Example} & \textbf{LLM Response} & \textbf{Explanation of errors} \\
\hline
Type 1: Reading or Interpretation Error  & 
\textit{Q: Using the information provided in \textbf{\textsc{Figure 5}}, which of the following is a claim that can be made? \newline 
A. Contractor is to purchase a unit manufactured by McQuay \newline 
B. The unit must have a minimum SEER of 7.5 \newline 
C. The unit has a cooling capacity under 1 ton \newline 
D. The roofing structure will not carry the dead weight of the unit \newline}  
\textbf{Correct:} D
&
\textbf{GPT-4o:} \textit{``A. Based on the Air Handling Unit Schedule shown in Figure 5, the correct answer is: \textbf{A. Contractor is to purchase a unit manufactured by McQuay.}"}
&
LLM misinterpreted the figure data, concluding that the contractor must buy a specific brand instead of recognizing the load limitation. This mistake stems from failing to extract the correct detail about structural capacity. \\

Type 2: Conceptual Misunderstanding &
\textit{Q: A \underline{\hspace{1cm}} happens at least once a month to review safety issues and contractor jobsite progress. \newline
A. Preconstruction conference \newline
B. Construction manager safety training \newline
C. Safety coordination meeting \newline
D. Safety committee meeting \newline}
\textbf{Correct:} D
&
\textbf{Claude 3.7:} \textit{``The correct answer is \textbf{c. Safety coordination meeting.} Safety coordination meetings are typically held at least once a month to review safety issues and contractor jobsite progress. These meetings bring together project stakeholders to discuss ongoing safety concerns ..."}
&
LLM recognized that there is a monthly safety meeting but used the wrong official term (“safety coordination meeting”) instead of the correct one (“safety committee meeting”). \\

Type 3: Procedural or Methodological Error &
\textit{Q: Using the bond premium schedule below, what is the bond premium for a \$2,755,000 project? First $\$5,000 = \$12.00/\$1,000$, Next $\$2M = \$6,000 + \$7.25/\$1,000$, Next $\$2.5M= \$20,500 + \$5.75/\$1,000$ \newline
A. \$19,974 \newline
B. \$45,373 \newline
C. \$55,974 \newline
D. \$66,341 \newline
}  
\textbf{Correct:} B
&
\textbf{GPT-4o:} \textit{``... Step 1: Break down the project amount into the tiers ... First $\$5,000:5\times 12=60$, Next $\$2,000,000: 2,000\times7.25=14,500$. Remaining $\$750,000: 750 \times 5.75 = 4,312.50$. Step 3: Add up the total premium. $60 +14,500+4,312.50=18,872.50$. Final Answer: \textbf{A. \$19,974} (The closest rounded answer, some rounding differences may exist, but this is the best match)."}
& LLM recognized the multi‐tier schedule but failed to carry out all steps properly. \\
\hline
\end{tabular}
\end{adjustbox}
\end{table}

\begin{figure}[htbp]
    \centering
    \includegraphics[width=0.8\linewidth]{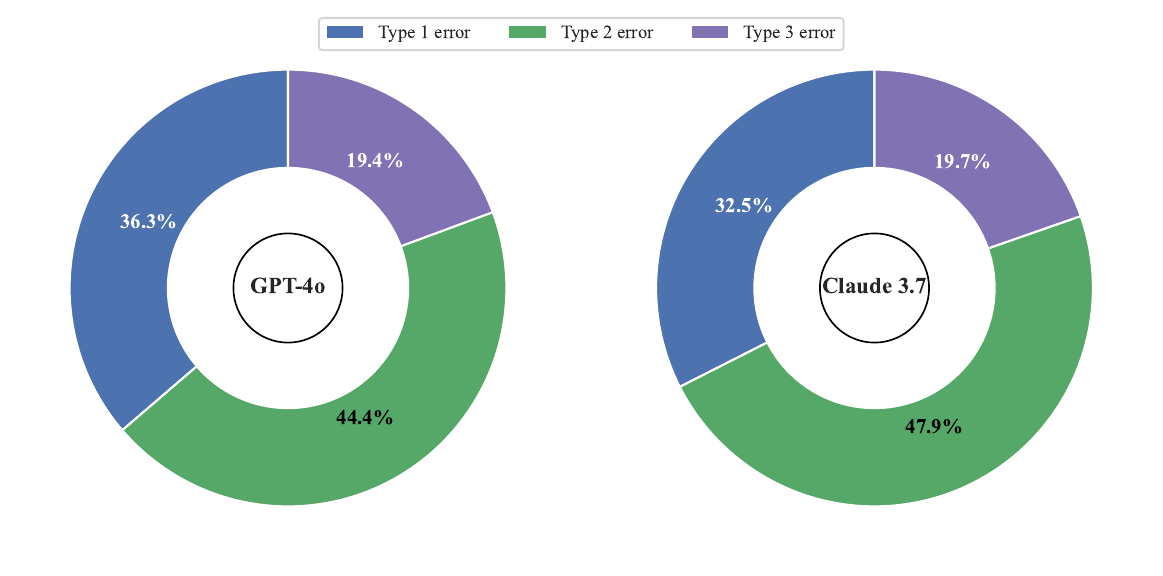}
    \vspace{-8mm}
    \caption{Distribution of error types for LLM responses}
    \label{fig:errors}
\end{figure}

The results in Fig.~\ref{fig:errors} indicate that GPT-4o and Claude 3.7 exhibit similar error distributions, with conceptual misunderstandings (Type 2 errors) being the most frequent. This suggests that while both models have strong language processing capabilities, they struggle with deeper domain-specific reasoning in CM tasks. A Chi-square test of independence (\(\chi^2=0.418, p=0.812\)) indicates that there is no statistically significant difference in error patterns between models, suggesting that these models share common failure modes, likely due to their reliance on general training data. This finding highlights the need for domain-adaptive fine-tuning or integration of external structured knowledge bases to improve their reasoning accuracy. Furthermore, reducing Type 1 and Type 3 failures could improve reliability in complex decision-making scenarios within CM applications.

\section{Discussion}
This section examines the implications of our findings in relation to human performance and broader applications in CM practice and education. We also discuss the limitations of current LLMs when applied to practical tasks and directions for future research.

\subsection{Benchmarking LLMs vs Human Performance}
\label{subsec:benchmark_llm_human}
Our comparative analysis demonstrates that state-of-the-art LLMs, including GPT-4o and Claude 3.7, can achieve accuracy levels exceeding 80\% on standardized multiple-choice assessments. This performance surpasses the 70\% passing threshold used for AIC-administered CM certification exams. For context, human performance on the CAC exam had only a 34\% pass rate among 691 candidates, underscoring the challenging nature of these assessments. Such high scores are likely attributable to extensive pre-training on diverse textual resources, which enables robust information retrieval and synthesis.

However, these high scores should not be equated with comprehensive professional competence, as there is no evidence that candidates who achieve significantly higher scores outperform those who score at the passing threshold in real-world CM practice \cite{CMAA2022}. The ability of LLMs to recall and reassemble textual information, while impressive, does not replicate the complex decision-making processes required in real-world project management. Human practitioners who pass CM certification exams also use substantial field experience and contextual judgment, qualities that LLMs inherently lack. The primary objective of this evaluation is not to determine whether LLMs can replace human expertise, but to assess their feasibility as supplementary tools for professional decision support in CM practices.

\subsection{Implications for Education and Industry Practice}
\label{subsec:implications_education_industry}
The robust performance of LLMs on CM examinations offers opportunities for academic and industry practice. In educational settings, models such as GPT-4o could serve as advanced tutoring aids by generating practice questions, providing detailed explanations of complex problem-solving strategies, and simulating real-world scenarios. Instructors can leverage these capabilities to design targeted assessments and interactive learning modules that foster critical thinking and active engagement. Early evidence in engineering education supports the idea that AI-driven tutoring, when properly supervised, can improve student understanding and facilitate deeper learning \cite{abril2024chatgpt}. However, we also need to note that over-reliance on LLMs may undermine learning outcomes if students rely solely on AI-generated solutions \cite{zhai2024effects}. Ultimately, LLMs should be integrated as supplemental resources that complement, rather than replace, hands-on practice and instructor-led instruction.

From an industry perspective, LLMs have the potential to streamline text-intensive tasks such as drafting requests for information, interpreting contract clauses, and generating safety protocols \cite{ghimire2023generative}. Assisting with such tasks can allow professionals to focus on higher-level strategic decision-making. However, it is critical to recognize the models’ limitations, particularly in handling complex multi-step reasoning and advanced quantitative tasks, which require careful oversight and validation by human experts before any operational deployment.

\subsection{Limitations of LLMs in Construction Management}
A detailed analysis of model output reveals that LLMs such as GPT-4o and Claude 3.7 demonstrate strong performance in qualitative and text-based reasoning. However, they face challenges in tasks requiring precise numeric computation and intricate multi-step problem-solving. For example, GPT-4o often provides detailed justifications for qualitative prompts but may struggle with calculations involving specialized formulas or figure-referenced information. Similarly, Claude can misinterpret subtle differences in terminology, leading to errors analogous to those seen in human practitioners with insufficient conceptual grounding.

In addition to technical shortcomings, ethical and legal considerations further complicate the integration of LLMs into CM workflows. The use of AI to process sensitive project data introduces potential confidentiality risks, and unresolved liability issues can arise if AI-generated recommendations are found to be incomplete or erroneous \cite{xiong2024transforming}. As a result, these factors underscore the need for robust human oversight and clearly defined accountability protocols to ensure that any AI assistance complements, rather than compromises, professional standards.

\subsection{Study Constraints and Future Work}
\label{subsec:study_constraints_future_work}
The scope of this study is limited by several factors. First, while the exam questions were curated from real certification materials, they may not fully capture the complexity and breadth of complete CM workflows. For example, we did not assess the models for generating comprehensive project schedules, performing detailed cost estimates, or drafting contract sections, tasks that are integral to CM practice. Additionally, it is important to note that success in an exam context, i.e., correctly answering questions, does not necessarily equal real-world competency, which also demands interpersonal skills, field experience, and adaptive judgment in dynamic scenarios \cite{barrows2020value}. Second, the dataset exhibits an under-representation of figure-referenced questions, with only 1.3\% of the total questions requiring figure interpretation (see Table \ref{tab:characteristics}). Many CM tasks, such as interpreting blueprints, schematics, and spatial layouts, rely on figure-based problem solving \cite{sacks2013building}. The dataset does not adequately assess these skills, creating a potential gap in our findings. Third, this study used a zero-shot prompting approach to evaluate LLMs' baseline capabilities in a single trial. Future research might explore adaptive prompting or domain-specific tuning to enhance performance \cite{sahoo2024systematic} and incorporate repeated evaluations to better assess consistency and reliability \cite{renze2024effect}. Finally, given the fast-paced advances in LLMs, newer models may outperform those evaluated in this study. Future improvements in reasoning, multi-modal processing, and domain adaptation could impact LLM performance in CM-related tasks, requiring continuous reassessment.

Despite these limitations, our study provides insight into the potential role of LLMs in CM practices. Future research should address these constraints by:
\begin{itemize}
    \item \textbf{Expanding practical task simulations:} Research could move beyond Q\&A assessments to evaluate LLM performance in practical CM tasks, such as developing project schedules, performing cost analyses, or drafting contractual documents. These tasks would offer a more comprehensive assessment of LLM utility in real-world settings.
    \item \textbf{Enhancing visual reasoning and figure interpretation:} Increasing the representation of figure-referenced questions and exploring multimodal AI models capable of integrating text, tables, and images will improve AI’s ability to interpret construction drawings, site plans, and engineering schematics.
    \item \textbf{Mitigating misconceptions and reasoning errors:} We observed that LLMs can sometimes produce reasoning errors or propagate common misconceptions. Future research should explore prompt engineering, verification steps, and hybrid AI models (e.g., human supervision) to improve accuracy in complex problem-solving for CM tasks.
    \item \textbf{Evaluating emerging LLMs and fine-tuning for CM applications:} As new models are released, repeated benchmarking will be valuable to track how LLM performance in CM exams improves over time. Additionally, fine-tuning LLMs using domain-specific corpora (e.g., project management textbooks, building codes, and historical data) could improve construction-specific reasoning and reduce domain errors.
    \item \textbf{Exploring Human-AI collaboration in CM education and practice:} Investigating AI-assisted decision-making in construction project teams, educational settings, and certification training could offer insights into how AI improves human expertise, reduces cognitive load, and supports industry professionals.
\end{itemize}

\section{Conclusions}
\label{sec:conclusions}
This work introduces \textsc{CMExamSet}, a curated dataset of CM certification exam questions designed to benchmark LLMs against professional standards. By compiling 689 MCQs from accredited CM certification materials, our dataset encapsulates a diverse range of knowledge areas, from cost control and contract administration to safety and ethics, varying levels of cognitive demand, including both single-step and multi-step reasoning, as well as question formats, including text-only, table-referenced, and figure-referenced.

Our evaluation of state-of-the-art LLMs, including GPT-4o and Claude 3.7, reveals that these models can achieve accuracy levels exceeding 80\% on exam-style questions, thereby surpassing the conventional 70\% passing threshold observed in human certification. This promising performance underscores the potential of LLMs to serve as supplementary decision-support tools in both CM education and practice. However, it is important to recognize that high exam performance does not equate to comprehensive professional competence. Unlike human practitioners, LLMs lack practical field experience, contextual judgment, and the ability to navigate the multifaceted nature of real-world project management.

The study also highlights critical limitations. Our evaluation was confined to exam-style Q\&A tasks, which do not fully represent the complex workflows of CM, such as project scheduling, detailed cost estimation, or contract drafting. Furthermore, the use of a zero-shot evaluation protocol, without using fine-tuning or interactive prompting, suggests that our findings reflect inherent model capabilities rather than optimized performance. Finally, the rapidly evolving nature of LLM technology requires a continuous reassessment as newer models emerge.

\section{Data Availability Statement}
\label{sec:data_availability}
The question sets in this study are from copyright-protected certification exams by the AIC and CMAA and cannot be publicly shared. Other data, models, or codes that support the findings of this study may be obtained from the corresponding author upon reasonable request.

\section{Acknowledgments}
This research was supported by startup funding from the Bert S. Turner Department of Construction Management at Louisiana State University (LSU) and Cajun Industries Professorship in Construction Management. The findings, interpretations, and conclusions expressed in this paper do not necessarily reflect the views of Cajun Industries or LSU.

\bibliography{ascexmpl-new}

\end{document}